\documentclass[sigconf]{acmart}

\usepackage{listings}

\AtBeginDocument{%
  }

\setcopyright{acmlicensed}
\copyrightyear{2026}
\acmYear{2026}
\setcopyright{cc}
\setcctype{by}
\acmConference[WWW '26]{Proceedings of the ACM Web Conference 2026}{April 13--17, 2026}{Dubai, United Arab Emirates}
\acmBooktitle{Proceedings of the ACM Web Conference 2026 (WWW '26), April 13--17, 2026, Dubai, United Arab Emirates}
\acmPrice{}
\acmDOI{10.1145/3774904.3792938}
\acmISBN{979-8-4007-2307-0/2026/04}

\begin{document}

\title{Truth with a Twist: The Rhetoric of Persuasion in Professional vs. Community-Authored Fact-Checks}

\author{Olesya Razuvayevskaya}
\orcid{0000-0002-7922-7982}
\email{o.razuvayevskaya@sheffield.ac.uk}

\affiliation{%
 \institution{The University of Sheffield}
  \city{Sheffield}
  \country{United Kingdom}
}

\author{Kalina Bontcheva}
\email{k.bontcheva@sheffield.ac.uk}
\orcid{0000-0001-6152-9600}
\affiliation{%
  \institution{The University of Sheffield}
  \city{Sheffield}
  \country{United Kingdom}}





\begin{abstract}

This study presents the first large-scale comparison of persuasion techniques present in crowd- versus professionally-written  debunks. Using extensive datasets from Community Notes (CNs), EUvsDisinfo, and the Database of Known Fakes (DBKF), we quantify the prevalence and types of persuasion techniques across these fact-checking ecosystems. Contrary to prior hypothesis that community-produced debunks rely more heavily on subjective or persuasive wording, we find no evidence that CNs contain a higher average number of persuasion techniques than professional fact-checks. We additionally identify systematic rhetorical differences between CNs and professional debunking efforts, reflecting differences in institutional norms and topical coverage. Finally, we examine how the crowd evaluates persuasive language in CNs and show that, although notes with more persuasive elements receive slightly higher overall helpfulness ratings,  crowd raters are effective at penalising the use of particular problematic rhetorical means.
\end{abstract}


\received[accepted]{13 January 2026}

\maketitle

\section{Introduction and Background}\label{sec:intro}

Community Notes (CNs) on X (formerly known as Twitter Birdwatch) is a crowd-driven content moderation mechanism for countering potentially misleading information by providing contextual clarifications, corrections, or additional sources \cite{Wirtschafter_Majumder_2023, de_keulenaar_content_2025}. This moderation process is self-sustaining: contributors both author notes and collectively evaluate their quality by voting on their helpfulness. Bridging algorithms subsequently determine which notes achieve sufficient cross-ideological agreement to be displayed publicly \cite{wojcik_birdwatch_2022, bouchaud_algorithmic_2025, ovadya_bridging_2023}. Prior research highlights the effectiveness of CNs in reducing the spread and impact of misinformation \cite{slaughter_community_2025, chuai_community-based_2024, drolsbach_diffusion_2023}, particularly for posts containing factual inaccuracies \cite{wang_efficiency_2024}. Such interventions frequently lead to post retractions by the original authors \cite{renault_collaboratively_2024, chuai_community-based_2024, gao_can_2024}, although they may also elicit increased user outrage or polarization in subsequent discussions \cite{chuai_community_2025}.

In comparison to professional fact-checking efforts, CNs offer broader coverage--both in the volume of misinformation addressed and in the diversity of topics requiring regional, cultural, or domain-specific expertise \cite{augenstein2025community}. Beyond these scalability advantages, users tend to express greater trust in corrective interventions when they are accompanied by CNs that explicitly articulate why a post is misleading, in contrast to opaque ``false information'' labels issued by experts \cite{drolsbach_community_2024}. Additionally, by engaging a substantially larger and more demographically diverse contributor base than traditional fact-checking organizations \cite{bouchaud_algorithmic_2025, augenstein2025community}, CNs promote a more democratic and transparent model of content governance. This inclusivity exposes readers to a wider range of perspectives and sources, particularly on controversial or geographically salient topics, reinforcing CNs’ potential as a scalable, community-centered approach to social-media accountability \cite{bouchaud_algorithmic_2025, augenstein2025community}.

Despite these strengths, there is no consensus among researchers on whether crowdsourced fact-checks are more susceptible to biases than professional ones \cite{augenstein2025community}. Some researchers argue that the bridging algorithm—which requires cross-ideological agreement for a note publication—reduces one-sidedness of CNs relative to professional fact-checks \cite{kangur_who_2024}. Others emphasise the system’s vulnerabilities to coordination and manipulation. For example, in the context of U.S. politics, posts by Republicans have been shown to receive more notes, both overall and published, than posts by Democrats \cite{doi:10.1073/pnas.2502053122}. Moreover, \citeauthor{10.1145/3531146.3534629} demonstrate that crowd workers’ assessments can diverge from expert-assigned ground-truth labels depending on their political orientations and broader cognitive or ideological dispositions \cite{10.1145/3531146.3534629}. They further find that crowd workers are more likely than experts to regard misleading claims as truthful \cite{10.1145/3531146.3534629}. Within this broader debate, some scholars argue that professional fact-checks are generally more objective in style, whereas CNs may employ persuasive or manipulative linguistic strategies, such as appeals to authority or emotionally charged claims \cite{Kankham18062025, augenstein2025community}.

Notwithstanding these concerns, no study to date has carried out a systematic comparison between professional and crowdsourced debunks with respect to their use of persuasive language. This gap motivates the present work, which aims to quantify the extent to which Community Notes adhere to their core values\footnote{https://communitynotes.x.com/guide/en/contributing/values} of avoiding personal opinions, insults, and manipulative behaviour. Accordingly, we pose the following research questions:
(a) Is the CNs' self-governing nature able to identify and penalise the use of persuasion techniques in community notes?
(b) Do community-written debunks exhibit a higher degree of persuasive language than professionally-authored ones?

Our code and supplementary material\footnote{\url{https://github.com/LesyaR/CN_Persuasion_Analysis}} are made openly available
to facilitate reproducibility.

\section{Datasets and Methodology}
This study utilises three datasets containing debunk texts: (1) Community Notes (CN), representing crowd-authored debunks, and two datasets representing debunks written by verification professionals, (2) EUvsDisinfo and (3) the Database of Known Fakes (DBKF). 
For the CN dataset, we extracted the full, daily updated set of notes on August 10, 2025, yielding 2,016,841 debunks. The EUvsDisinfo dataset consists of professionally-written debunks addressing pro-Kremlin disinformation \cite{leite2024euvsdisinfo}. Our experiments use only the debunk articles rather than the associated misinformation ones. Since a single article may debunk multiple misinformation claims, only unique articles are retained, resulting in 8,522 debunks in English.
The DBKF database\footnote{https://dbkf.ontotext.com/}
 contains a large, diverse collection of 681,346 debunks published by fact-checking organisations worldwide, which makes it more directly comparable in scale to CNs. 
 Unlike the EUvsDisinfo debunks which are English only, the DBKF and CNs datasets are highly multilingual, containing debunks in 56 and 72 languages respectively. The language distributions for CNs and DBKF are provided in the supplementary material.

Persuasion techniques in debunk texts are detected using the state-of-the-art multilingual classifier developed by Wu et al. \cite{wu-etal-2023-sheffieldveraai}. To increase confidence in individual predictions, the classification threshold is set to 0.8. For consistency across datasets and due to the limitations of transformer-based models, all texts were truncated to a maximum length of 512 tokens. The full list of techniques, along with their definitions, is provided in the supplementary material.

\textbf{RQ1. Is the
CNs’ self-governing nature able to identify and penalise the use of persuasion techniques in community notes?}
To address this question, we examined two aspects. First, using the complete set of note ratings from the CN website, we assessed whether the \emph{degree of note persuasiveness} is correlated with the \emph{degree of note helpfulness}. We operationalised the degree of note persuasiveness as the number of persuasion techniques identified in a given note. To estimate the helpfulness degree, we considered individual user ratings that mark the note as being \textit{Not Helpful}, \textit{Somewhat Helpful}, or \textit{Helpful}. We then mapped each note rating to a numeric value: \textit{Not Helpful}->1, \textit{Somewhat Helpful}->2, or \textit{Helpful}->3. For notes with multiple ratings, we calculated the average rating value. The association between the degree of persuasiveness and helpfulness level is measured using the Spearman correlation coefficient \cite{wissler1905spearman}. For each persuasion technique, we additionally compared the distribution of helpfulness scores between notes that use the technique and those that do not, using the Mann--Whitney U test \cite{test2000mann}. To assess the \emph{direction} and \emph{magnitude} of the effect, we computed Cliff's delta (\(\delta\)), a non-parametric effect size metric \cite{macbeth2011cliff}.

\textbf{RQ2.  Do community-
written debunks exhibit a higher degree of persuasive language
than professionally-authored ones?}
To compare CN with the two professional fact-checking datasets, we first removed CN notes that classify posts as “Not Misinformation”, ensuring comparability with EUvsDisinfo and DBKF, which publish debunks only for misinformation claims. We then performed two analyses. First, we tested whether the CNs dataset contains more persuasion techniques per note on average compared to professionally-written debunks, EUvsDisinfo and DBKF. Given the binary nature of the data and large sample sizes, we used the non-parametric Mann-Whitney U test to compare distributions between datasets \cite{test2000mann}.

We next performed a per-technique comparison by formulating the following directional hypotheses for each technique:
\begin{itemize}
    \item \textbf{$H_{1}$:} The prevalence of a given technique is higher in CNs than in DBKF.
    \item \textbf{$H_{2}$:} The prevalence of a given technique is higher in CNs than in EUvsDisinfo.
\end{itemize}
For each comparison, the $H_{0}$ hypothesis assumes that CNs have equal or lower prevalence for each technique. Because the data for each technique are binary (presence/absence), we employed a one-sided \textit{proportions z-test} to test the directional hypotheses \cite{webb2017mostly}.

\section{Results}
 
\noindent{\textbf{RQ1. Is the CNs' self-governing nature able to identify and penalise the use of persuasion techniques in community notes?}
} We first evaluated whether the degree of persuasiveness of notes was correlated with their average helpfulness rating. Our results show the correlation to be statistically significant, but the effect size is extremely small ($\rho \approx 0.039$, $p$-value$<10^{-4}$). Importantly, the direction of correlation is \emph{positive}, indicating that, contrary to our hypothesis, notes containing more persuasion techniques are, on average, rated \emph{more} helpful by the crowd. 

Table~\ref{tab:directional-deltas} reports the \(p\)-values and Cliff’s deltas estimated based on Mann--Whitney U statistic for each technique. Positive $\delta$ values indicate that the technique is associated with \emph{higher} helpfulness, while negative values indicate the \textit{lower} helpfulness association and near-zero numbers indicate small effect, often minimal in practice. As can be seen, several techniques exhibit small but meaningful directional effects. The largest positive effects were observed for \textit{Consequential Oversimplification} (\(\delta = 0.255\)), \textit{Causal Oversimplification} (\(\delta = 0.115\)), \textit{Appeal to Hypocrisy} (\(\delta = 0.114\)), and \textit{Appeal to Time} (\(\delta = 0.1053\)), indicating that notes containing these techniques tend to receive \emph{higher} helpfulness ratings.
Conversely, techniques such as \textit{Flag Waving} (\(\delta = -0.142\)), \textit{Guilt by Association} (\(\delta = -0.088\)), and \textit{Straw Man} (\(\delta = -0.068\)) were associated with \emph{lower} helpfulness ratings.
Most remaining techniques showed very small effects (\(|\delta| < 0.05\)), suggesting limited practical impact on helpfulness despite highly significant \(p\)-values driven by large sample size. Two techniques, \textit{Appeal to Authority} and \textit{False Dilemma / No Choice}, were not statistically significant and exhibited negligible effect sizes, indicating no detectable relationship with helpfulness. For \textit{Appeal to Authority}, this can be attributed to its general association with trustworthy narration and objective presentation \cite{leite2025weakly}.
Overall, whereas the degree of persuasiveness has a minimal directional effect, individual techniques do exhibit systematic positive or negative associations with perceived helpfulness.

\begin{table}[h!]
\centering
\caption{Directional effects of individual persuasion techniques on helpfulness (Mann--Whitney U tests).}
\label{tab:directional-deltas}
\begin{tabular}{lrr}
\hline
\textbf{Technique} & \textbf{\(p\)} & \textbf{\(\delta\)} \\
\hline
Appeal to Authority                 & 0.064               & -0.005 \\
Appeal to Fear-Prejudice            & $3\times10^{-5}$    & -0.010 \\
Appeal to Hypocrisy                 & $<1\times10^{-300}$ & 0.114 \\
Appeal to Popularity                & $1\times10^{-80}$   & 0.066 \\
Appeal to Time                      & $1\times10^{-9}$    & 0.105 \\
Appeal to Values                    & $2\times10^{-64}$   & 0.051 \\
Causal Oversimplification           & $7\times10^{-267}$  & 0.115 \\
Consequential Oversimplification    & $<1\times10^{-300}$ & 0.255 \\
Conversation Killer                 & $3\times10^{-35}$   & -0.025 \\
Doubt                               & $1\times10^{-57}$   & 0.017 \\
Exaggeration-Minimisation           & $8\times10^{-222}$  & 0.061 \\
False Dilemma-No Choice             & 0.227               & 0.005 \\
Flag Waving                         & $<1\times10^{-300}$ & -0.142 \\
Guilt by Association                & $3\times10^{-111}$  & -0.088 \\
Loaded Language                     & $<1\times10^{-300}$ & 0.041 \\
Name Calling-Labeling               & $1\times10^{-102}$  & 0.027 \\
Obfuscation-Vagueness               & $4\times10^{-185}$  & 0.085 \\
Questioning the Reputation          & $8\times10^{-20}$   & -0.023 \\
Red Herring                         & $1\times10^{-17}$   & 0.042 \\
Repetition                          & $<1\times10^{-300}$ & 0.079 \\
Slogans                             & $8\times10^{-15}$   & -0.032 \\
Straw Man                           & $1\times10^{-11}$   & -0.068 \\
Whataboutism                        & $1\times10^{-44}$   & 0.094 \\
\hline
\end{tabular}
\end{table}

\textbf{RQ2. Do community-
written debunks exhibit a higher degree of persuasive language
than professionally-authored ones?}
We first tested  whether the CNs dataset contains more persuasion techniques on average compared to the EUvsDS and DBKF datasets using the Mann-Whitney U test. The results for both comparisons were insignificant, with p-value=1.00. This suggests that contrary to prior research (Section~\ref{sec:intro}), it is not the case that the argumentation of crowd-written fact-checks is more persuasive in nature as compared to professionally-written fact-checks.

Table~\ref{tab:per_technique} summarises the results of the one-sided z-tests for per-technique presence comparison between the datasets. Positive z-values indicate higher prevalence in CNs, and p-values below 0.05 indicate statistically significant differences. The results indicate that the prevalence of persuasion techniques in CNs (underlined values) differs markedly depending on the comparison dataset. 
Compared to DBKF, only a few techniques (\textit{Conversation Killer}, \textit{Repetition}, \textit{False Dilemma-No Choice}, and \textit{Whataboutism}) are significantly more prevalent in CNs. For most other techniques, prevalence is lower in CNs, contrary to our initial hypothesis. Compared to EUvsDisinfo, much more techniques (\textit{Appeal to Hypocrisy}, \textit{Appeal to Popularity}, \textit{Appeal to Values}, \textit{Consequential Oversimplification}, \textit{Conversation Killer}, \textit{Doubt}, \textit{Obfuscation-Vagueness-Confusion}, \textit{Questioning the Reputation}, \textit{Red Herring}, \textit{Straw Man}, and \textit{Whataboutism}) are significantly more prevalent in CNs. This can be attributed to the stricter format  and the domain specificity of the EUvsDisinfo debunks.

\begin{table*}[h!] 
\centering
\caption{Per-technique one-sided z-tests comparing prevalence in \textit{CNs} vs \textit{DBKF} and \textit{EUvsDisinfo}.}
\label{tab:per_technique}
\begin{tabular}{lrr|rr}
\hline
{\textbf{Technique}} & \multicolumn{2}{c|}{\textbf{CNs vs DBKF}} & \multicolumn{2}{c}{\textbf{CNs vs EUvsDisinfo}} \\
\cline{2-5}
 & Z-statistic & p-value & Z-statistic & p-value \\
\hline
Appeal to Authority & -124.10 & 1.000 & -44.70 & 1.000 \\
Appeal to Fear-Prejudice & -160.98 & 1.000 & -86.84 & 1.000 \\
Appeal to Hypocrisy & -128.00 & 1.000 & \underline{16.04} & $<0.001$ \\
Appeal to Popularity & -73.97 & 1.000 & \underline{5.18} & $<0.001$ \\
Appeal to Time & -50.45 & 1.000 & \underline{2.20} & 0.014 \\
Appeal to Values & -107.68 & 1.000 & \underline{12.85} & $<0.001$ \\
Causal Oversimplification & -18.30 & 1.000 & -48.28 & 1.000 \\
Consequential Oversimplification & -48.95 & 1.000 & \underline{6.22} & $<0.001$ \\
Conversation Killer & \underline{45.36} & $<0.001$ & \underline{19.08} & $<0.001$ \\
Doubt & -187.06 & 1.000 & \underline{3.97} & 0.00004 \\
Exaggeration-Minimisation & -120.47 & 1.000 & -17.01 & 1.000 \\
False Dilemma-No Choice & \underline{22.85} & $<0.001$ & -0.82 & 0.79 \\
Flag Waving & -88.85 & 1.000 & -60.16 & 1.000 \\
Guilt by Association & -111.78 & 1.000 & -61.53 & 1.000 \\
Loaded Language & -84.67 & 1.000 & -18.40 & 1.000 \\
Name Calling-Labeling & -205.16 & 1.000 & -79.70 & 1.000 \\
Obfuscation-Vagueness-Confusion & -5.62 & 1.000 & \underline{12.04} & $<0.001$ \\
Questioning the Reputation & -267.89 & 1.000 & \underline{15.41} & $<0.001$ \\
Red Herring & -76.88 & 1.000 & \underline{4.84} & $<0.001$ \\
Repetition & \underline{137.26} & $<0.001$ & -21.62 & 1.000 \\
Slogans & -244.23 & 1.000 & -1.05 & 0.85 \\
Straw Man & -77.73 & 1.000 & \underline{2.14} & 0.016 \\
Whataboutism & \underline{6.98} & $<0.001$ & \underline{4.86} & $<0.001$ \\
\hline
\end{tabular}
\end{table*}

\section{Discussion and Conclusions}

This paper reported the first large-scale examination of the widely held assumption that crowdsourced debunks tend to employ more persuasive wording than professional ones. Contrary to concerns that the decentralised, volunteer-driven nature of CNs might make them less objective, we found no evidence that CNs employ a greater number of persuasion techniques than professional fact-checks. 

It should be noted, however, that our per-technique comparisons reveal a more complex picture. While CNs are less persuasive than the DBKF fact-checks along most dimensions, they show higher prevalence for several techniques when compared to EUvsDisinfo. This highlights the importance of domain and organisational context: EUvsDisinfo debunks are produced within a niche geopolitical domain with stringent editorial oversight, whereas DBKF addresses a far broader thematic and organisational space. The differences therefore underline that rhetorical variation is shaped not only by authorship (crowd vs. professional) but also by institutional aims, editorial norms, and the type of misinformation being countered.

The analysis of rater behaviour further extends existing knowledge of CNs’ self-regulating mechanisms. The overall degree of note persuasiveness exhibits only a negligible relationship with its average helpfulness degree, and—contrary to our hypothesis—the association is slightly positive. 
However, examining the presence of individual techniques reveals that crowd raters are particularly sensitive and penalise certain types of persuasive rhetoric, especially patriotically charged claims (\textit{Flag-Waving}); statements that exhibit, biases against individuals through association with specific groups (ethnic, racial, political, ideological, etc.; \textit{Guilt by Association}); and the notes that purposefully misrepresent the original post by addressing an argument that was not made by the post author (\textit{Straw Man}). 
The latter crowd reaction could represent a particular case of the widely observed NNN (Note Not Needed) flagging phenomenon in CNs, whereby users label notes as irrelevant to the post \cite{razuvayevskaya2025timeliness}.
In contrast, the use of other techniques appears to persuade the raters and increase their trust in CNs. This suggests that CN contributors are sensitive to specific problematic rhetorical strategies even if they do not penalise their cumulative presence.

\section{Acknowledgments}

This  work  has  been  co-funded  by  the  UK’s innovation agency grant 10039055 (approved under the Horizon Europe as vera.ai,  EU  grant  agreement  101070093) under action number 2020-EU-IA-0282.


\bibliographystyle{ACM-Reference-Format}


\end{document}